\documentclass[twoside,leqno,twocolumn]{article}

\usepackage[letterpaper]{geometry}
\usepackage{ltexpprt}
\usepackage{hyperref}
\usepackage{graphicx}
\usepackage{subcaption}
\usepackage{enumitem}
\usepackage{amsmath}
\usepackage{breqn}
\usepackage{lipsum}
\usepackage{multirow}
\usepackage[normalem]{ulem}
\usepackage{caption}
\usepackage{tabu}
\usepackage{multibib}
\usepackage{algorithm}
\usepackage{algpseudocode}
\usepackage[T1]{fontenc}
\usepackage{cancel}
\usepackage{xcolor}
\usepackage{titlesec}
\usepackage{balance}

\usepackage{array}
\usepackage{adjustbox}
\usepackage{mathtools}
\usepackage{cellspace}
\usepackage{float}
\usepackage{mathrsfs}

\usepackage[noadjust]{cite}

\newtheorem{definition}{Definition}

\usepackage{etoolbox}
\usepackage{titlesec}
\begin{document}









\title{Spatially-Delineated Domain-Adapted AI Classification: An Application for Oncology Data}



\author{%
Majid Farhadloo\(^{\dagger}\), Arun Sharma\(^{\dagger}\), Alexey Leontovich\(^{\S}\), Svetomir N. Markovic\(^{\S}\), Shashi Shekhar\(^{\dagger}\) \\
\(^{\dagger}\) \textit{Department of Computer Science \& Engineering, University of Minnesota} \\
\(^{\S}\) \textit{Department of Oncology, Mayo Clinic} \\
Emails: \(^{\dagger}\)\{farha043, sharm485, shekhar\}@umn.edu,\\ 
\(^{\S}\)\{leontovich.alexey, markovic.svetomir\}@mayo.edu
}


\date{}

\maketitle

\fancyfoot[R]{\scriptsize{Copyright \textcopyright\ 2025 by SIAM\\
Unauthorized reproduction of this article is prohibited}}

\begin{abstract} \small\baselineskip=11pt
Given multi-type point maps from different place-types (e.g., tumor regions), our objective is to develop a classifier trained on the source place-type to accurately distinguish between two classes of the target place-type based on their point arrangements. This problem is societally important for many applications, such as generating clinical hypotheses for designing new immunotherapies for cancer treatment. The challenge lies in the spatial variability, the inherent heterogeneity and variation observed in spatial properties or arrangements across different locations (i.e., place-types). Previous techniques focus on self-supervised tasks to learn domain-invariant features and mitigate domain differences; however, they often neglect the underlying spatial arrangements among data points, leading to significant discrepancies across different place-types. We explore a novel multi-task self-learning framework that targets spatial arrangements, such as spatial mix-up masking and spatial contrastive predictive coding, for spatially-delineated domain-adapted AI classification. Experimental results on real-world datasets (e.g., oncology data) show that the proposed framework provides higher prediction accuracy than baseline methods.\\
\textit{\textbf{Keywords:} spatially-delineated domain-adapted AI classifier, spatial arrangements, transfer learning, oncology}
\end{abstract}

\maketitle

\section{Introduction}
Spatial variability is a prominent feature exhibited by many phenomena. In ecology, spatial variability influences species distribution and ecosystem processes such that different plant species might dominate specific areas due to variations in soil quality, light availability, or moisture levels \cite{turner2005causes}. In agriculture, to account for spatial variability is essential for optimizing resource management and interventions to maximize crop yield \cite{paccioretti2020fastmapping}. In medical sciences, spatial variability highlights the interaction between maternal and fetal cells across the placental barrier, a crucial factor in Rh incompatibility where an Rh-negative mother may develop an immune response against her Rh-positive fetus \cite{moise2008management}. Therefore, comprehending spatial variability is crucial for understanding spatial patterns and phenomena in various contexts. Section \ref{application_domain} describes a bio-medical application domain showcasing the significance of spatial variability in oncology. 

 
Given multi-type point maps (e.g., cellular maps) from different place-types (e.g., tumor regions such as tumor-core or interface), our objective is to develop a classifier that can accurately distinguish between two classes (e.g., responder and non-responder) of the target place-type based on their point arrangements. This classifier will be trained using only source place-type learning samples and unlabeled target place-type instances and primarily relies on self-supervision of the data itself. This problem, referred to as unsupervised domain adaptation, involves classifying data from a target place-type distribution using only labeled samples from a different, source place-type distribution.

\begin{figure}[h]
    \centering
    \includegraphics[width=0.9\linewidth]{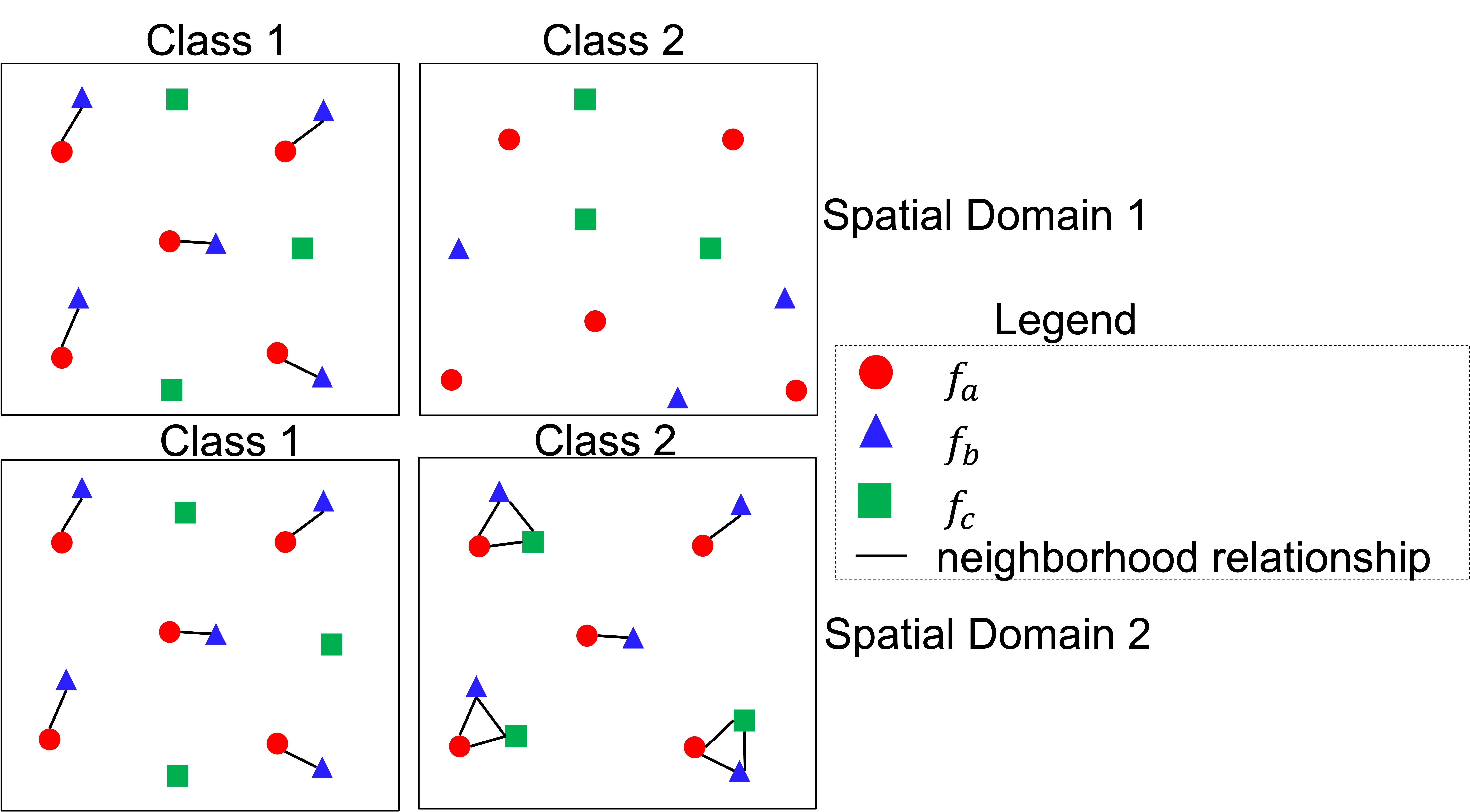}
    \caption{Spatial differences in arrangements between two class labels of multi-type point maps from different spatial domains.}
    \label{input_output}
\end{figure}

However, this problem is challenging due to the following reasons. First, spatial variability presents a significant challenge, wherein includes variations in distributions, density, and other morphological features among different place-types. Patterns and arrangements indicative of a class in one domain might not apply in another. As shown in Fig. \ref{input_output}, Spatial Domain $1$ distinguishes between class $1$ and class $2$ using the arrangements of <\textcolor{red}{circle} and \textcolor{blue}{triangle}>. However, due to spatial variability, this same pattern does not distinguish class labels in Spatial Domain $2$, where the distinct arrangement is a three-way relationship among the <\textcolor{red}{circle}, \textcolor{blue}{triangle}, and \textcolor{green}{square}> data points. A second challenge is that the success of deep neural networks relies on the availability of large labeled datasets, which are time-consuming and expensive to obtain, especially in medical sciences where domain experts are required for annotations \cite{willemink2020preparing}. 

Domain adaptation has become an emerging area of research to address these challenges. Most prior studies on domain adaptation for spatially delineated AI classification of multi-type point maps rely on adversarial methods \cite{ganin2016domain, NIPS2019_8940, debortoli2021adversarial, saito2018maximum, xu2020adversarial} or self-supervised learning approaches \cite{achituve2021self, shen2022domain, zou2021geometry, Liu2023}. Domain adversarial neural networks \cite{ganin2016domain} use a gradient reversal layer (GRL) to encourage the feature extractor to produce more similar features across domains as the domain classifier tries to differentiate between source and target features. PointDAN \cite{NIPS2019_8940} addresses point set classification via domain adaptation, using adversarial training with maximum classifier discrepancy \cite{saito2018maximum}, along with a GRL, which jointly learns and aligns both local and global features across source and target domains. On the other hand, self-supervised learning techniques designs tasks that do not require labeled data but instead leverage the inherent structure of the data to learn domain-invariant features and mitigate domain differences. Achituve et al. \cite{achituve2021self} enhance model training by introducing volume-based and sample-based deformation reconstruction for shape prediction, and point cloud mix-up to estimate proportions of mixed inputs. GAST \cite{zou2021geometry} introduces a geometry-aware, self-supervised training method that encodes domain-invariant geometric features into semantic representations to mitigate domain discrepancies in point-based representations. However, these works overlook the underlying spatial arrangements among data points, leading to substantial discrepancies among different place types. 

To address this limitation, we explore a novel multi-task architecture where self-supervised tasks are defined to learn representations that capture the underlying spatial arrangements of the target place-type. These tasks involve spatial mix-up masking of sub-regions across place-types to classify and predict their proportions, while ensuring the model captures spatially consistent features by contrasting positive and negative pairs,  which includes both location-dependent and independent arrangements.

\textbf{Our contributions are as follows:}
\begin{itemize}
    \item We propose a spatially-delineated domain-adapted AI classification model within a multi-task architecture that focuses on spatial arrangements through self-supervised learning to enhance unsupervised domain adaptation for multi-type point maps.
    \item Experiments show the proposed model outperforms existing baseline methods.
    \item A case study highlights the impact of spatial variability on tumor classification. The aim is to enhance the manual assessments provided by pathologists.
\end{itemize}

\textbf{Scope:} 
This paper explores spatial-interaction and variability-aware AI for multi-type point maps, focusing on spatially-delineated domain-adapted classification. Since these data function as permutation-invariant sets without predefined connectivity, comparisons with graph CNNs \cite{zhang2019graph} are excluded. 

\textbf{Organization:} The rest of the paper is organized as follows.  Section \ref{application_domain} briefly describes an application domain of this problem. Section \ref{sec:Problem} introduces key concepts and formally defines the problem. Our proposed methods are described in Section \ref{sec:ProposedApproach}. Section \ref{sec:Experiment} presents the evaluation of the proposed method, followed by a case study in Section \ref{case-study}. Related work is reviewed in Section \ref{related_work}. Section \ref{sec:Conclusion} concludes the paper and outlines future work.
\section{An Illustrative Application
Domain}\label{application_domain}
In cancer research, understanding tumor interactions with normal tissues is essential for insights on disease progression and immune therapy development. Multiplexed immunofluorescence (MxIF) imaging, especially relevant for immunocheckpoint inhibitor therapy (ICI), offers a detailed cellular spatial map of tumor and immune cells. Fig. \ref{fovs}, shows a map of different cell types (e.g., tumor and immune cells) with their corresponding locations. Although ICI therapy targets cancer cells by activating specific T lymphocytes, its effectiveness depends on complex \textit{spatial arrangements} within the tumor microenvironment (TME) \cite{andreou2022multiplexed, li2022cscd}.

While recent studies \cite{farhadloo2024towards, farhadloo2024spatial, nawaz2016computational} offer insight into TME heterogeneity, modeling its spatial variability is challenging due to factors such as rapid cancer cell proliferation, genetic instability, and the presence of unknown mediator cells with immune and target cells. Figure \ref{h&e} displays spatial variability across a tissue slide using three colored anatomic structures. Furthermore, the MxIF workflow can sometimes result in compromised data quality due to factors such as tissue loss because of iterative cycles of slide washing/staining. Therefore, by leveraging surrounding anatomical structures within remaining tissue on the slide, our method algorithmically characterizes spatial patterns to enhance pathologists' visual assessments and compensate for compromised areas on the slide, thereby mitigating the impact of poor-quality data.

\section{Problem Formulation} \label{sec:Problem}
This section reviews key concepts related to our work and presents the formal problem statement.

\subsection{Basic concepts}
\begin{figure}[h]
    \centering
    \includegraphics[width= \linewidth]{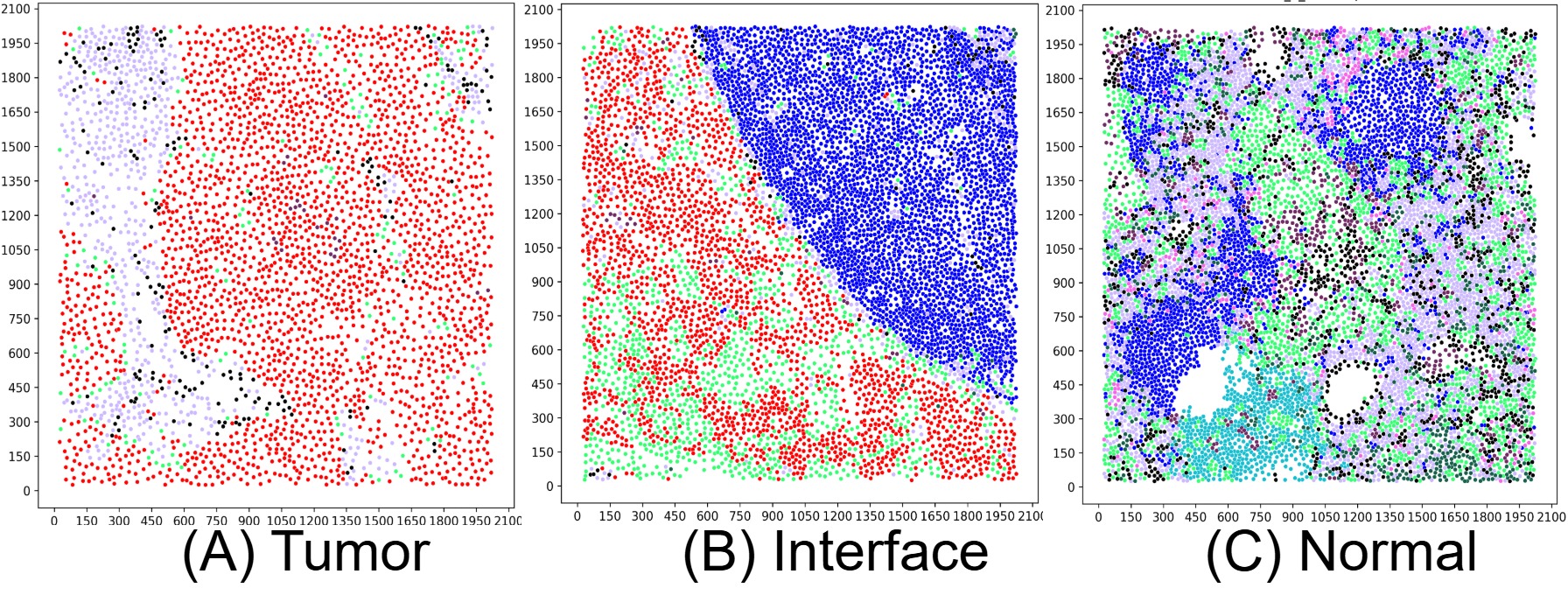}
    
    \caption{Three multi-type point maps are categorized based on tumor infiltration. }
    \label{fovs}
\end{figure}

\begin{definition}

A \textbf{place-type} is a spatial domain $\mathscr{X}$ associated with a probability distribution $P(X)$ over instances $X = \{p_i = (l_i, c_i) | p_i \in \mathscr{X}, i = 1, ..., n\}$, where $c_i$ is a non-spatial categorical attribute, $l_i$ is a two-dimensional vector of spatial point features. The set $X$ forms a \textbf{multi-type point map} representing objects (e.g., different cell types) and locations.
\end{definition}
For instance, Fig. \ref{fovs} displays three multi-type point maps from different place types (i.e., tumor, interface, and normal) classified based on tumor infiltration and cell population.

\begin{definition}
A \textbf{spatial arrangement} is the relative positions, or orientations of spatial objects in a space, as well as the patterns and relationships that emerge from their arrangement. 
\end{definition}
For example, co-location patterns \cite{shekhar2001discovering, intro_SDM} are spatial arrangement where subsets of objects frequently occur in close proximity.

\begin{figure}[h]
    \centering
    \includegraphics[width= \linewidth]{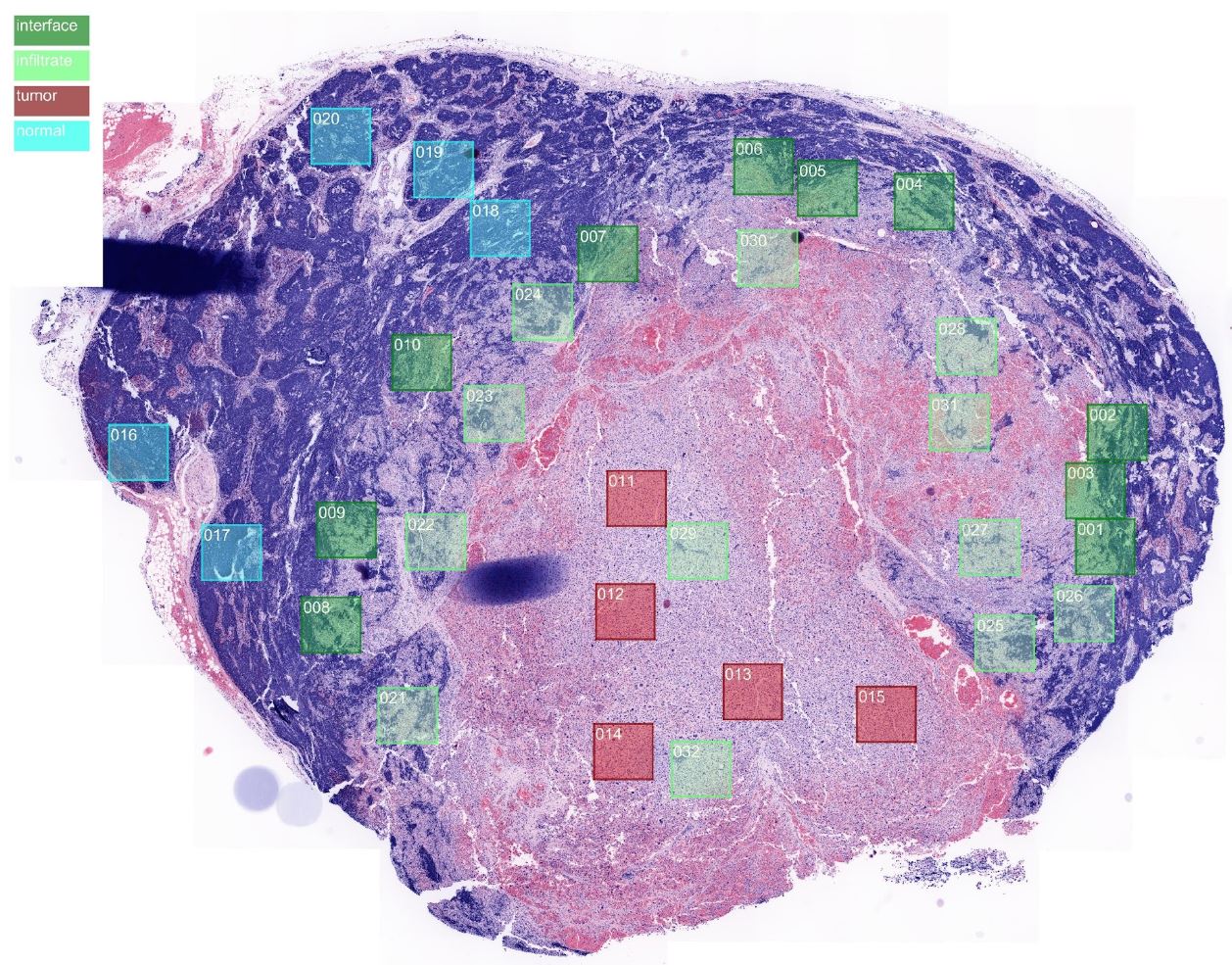}
    \caption{Spatial variability in a tissue slide, emphasized by pathology-driven fields of view (FOV, colored squares), reveals distinct patterns within each FOV.}
    \label{h&e}
\end{figure}

\begin{definition}
\textbf{Spatial variability} refers to the inherent heterogeneity and variation observed in a set of spatial patterns, structures, properties or arrangements across spatial domains (i.e., place-types).
\end{definition}
For example, in Fig. \ref{fovs}, the three selected multi-type point maps (from Fig. \ref{h&e}) depicting place-types (e.g., tumor, interface, and normal regions) illustrate significant spatial variability, revealing variations in cell population across a single tissue sample.

\begin{definition}
A \textbf{spatially-delineated domain adapted classification} $C$ consists of a labeled source place-type $PT_S = {(X_i^s, Y_i^s)}_{i=1}^{n_s}$, a target place-type $PT_T = {(X_i^t)}_{i=1}^{n_t}$, and a decision function $f_A$ that incorporates underlying \textbf{spatial arrangements $A$} among data points of source and target place-types by modeling \textbf{spatial variability} to address inherent heterogeneity in spatial patterns across place types. The decision function $f_A$ is defined as $f_A(x_i^t, x_i^s, a_i^s, a_i^t) = {P(y_k|x_i^t, x_i^s, a_i^s, a_i^t) | y_k \in \mathcal{Y}, k = 1, ..., |\mathcal{Y}|}$, where $x_i^t$ and $a_i^t$ represent an instance and corresponding arrangements from target place-type $PT_T$, and similarly, $x_i^s$ and $a_i^s$ are an instance and arrangements from source place-type $PT_S$. The decision function maps instances $x_i^t$ to label probabilities based on arrangements and variability observed across place-types.
\end{definition}

\subsection{Problem Statement}: The problem of spatially-delineated domain-adapted unsupervised AI classification of multi-type point maps can be expressed as follows:\\
\textbf{Input:}
\setlist{nolistsep}
\begin{itemize}
    \item [--] A labeled source place-type $PT_S = \{(X_i^s, Y_i^s)\}_{i=1}^{n_s}$ 
    \item [--] A target place-type $PT_T = \{(X_i^t)\}_{i=1}^{n_t}$  and 
    \item [--] $ Y_i^s \in \mathcal{Y} = \{1, \ldots, Y\},$ where $\mathcal{Y}_S = \mathcal{Y}_T$
\end{itemize}
\textbf{Output:} A domain-adapted classifier \( C \) for class separation within target place-type \( PT_T \).\\
\textbf{Objective:}  Solution quality (e.g., Accuracy, F1-score)
\textbf{Constraints:} 
\begin{itemize}
 \item [--] Spatial variability,
 \item [--] unlabeled target place-type instances
\end{itemize}

In this problem, we have access to multi-type point maps from a labeled source place-type \( PT_S = \{(X_i^s, Y_i^s)\} \) and an unlabeled target place-type \( PT_T = \{(X_i^t)\} \), where each multi-type map is associated with one of two class labels (e.g., responder and non-responder). We assume that the target place-type shares the same label space as the source place-type, denoted \( \mathcal{Y}_S = \mathcal{Y}_T \). Both the source place-type \( PT_S \) and the target place-type \( PT_T \) are associated with distinct probability distributions \( P_S(X) \) and \( P_T(X) \), respectively. Due to spatial variability, the identically distributed assumption is violated, implying that \( P_S(X) \neq P_T(X) \). The objective of the spatially-delineated domain-adapted AI classifier for unsupervised domain adaptation is to accurately classify target place-type instances based on their spatial arrangements. 

\begin{figure*}
    \centering
    \includegraphics[width=0.7\linewidth]{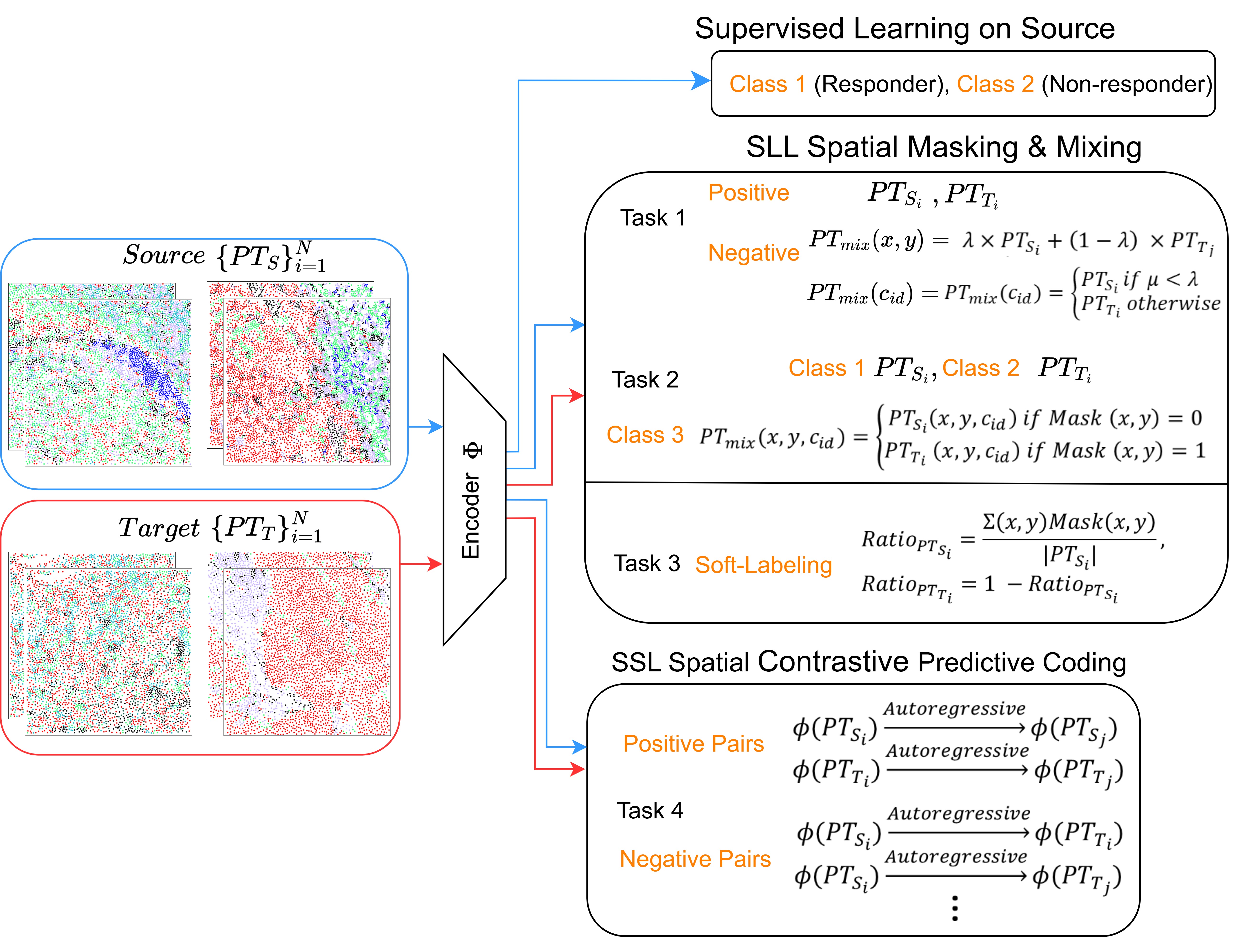}
    \caption{The overall framework of the proposed work.}
    \label{Proposed}
\end{figure*}

The overall proposed approach is illustrated in Fig. \ref{Proposed}. The domain-adapted classifier is integrated within a multi-task learning framework, employing supervised loss on source place-type instances along with spatially-oriented self-supervised learning (SSL) tasks that encompass both source and target place-types to enhance domain adaptation. This includes self-supervised spatial masking and mixing techniques, which expose the model to diverse spatial arrangements and make it robust to noise. Additionally, spatial contrastive predictive coding is proposed to discern both location-dependent and location-independent spatial arrangements across the source and target place-types.

\section{Proposed Approach}\label{sec:ProposedApproach}
In this section, we describe the proposed spatially-delineated domain-adapted AI classifier, which consists of spatially-oriented self-supervised learning tasks divided into two sub-modules: spatial mix-up masking and spatial contrastive predictive coding. To account for spatial variability, we explore various pseudo ground truths to train the network using both source and target instances for unsupervised domain adaptation.

\subsection{Spatial Mix-up Masking} \label{spatialMM}: 
Spatial mix-up masking extends the traditional mix-up technique \cite{zhang2017mixup} by applying it to both spatial and non-spatial categorical features of multi-type point maps. This method creates new synthetic samples that serve as spatial intermediates between source and target place-type instances. We define a set of self-supervised learning tasks within the family of spatial mix-up masking. These tasks expose the model to handle a broader range of variations, making it less sensitive to specific place-type characteristics and more attuned to underlying spatial arrangements. The learning tasks also enable the model to adapt to changing data distributions and to learn progressively. The objective focus of these tasks is to discriminate between real (source or target) and mix-up samples.

\subsubsection{Spatial Mix-up}: For each pair of source and target instances, $p_i^s = (l_i^s = (p_x^s, p_y^s), c_i^s) \in X_S$ and $p_i^T = (l_i^t = (p_x^t, p_y^t), c_i^t) \in X_T$, the mixed spatial and non-spatial categorical attributes are computed as follows: 
\begin{enumerate}
    \item \textbf{Mixed spatial attributes}:
    \begin{equation}
    l_{\text{mix}_i} = \lambda \cdot (p_x^s, p_y^s) + (1- \lambda) \cdot (p_x^t, p_y^t),\\
    \lambda \sim \beta(\alpha, \alpha)
    \end{equation}
    \item \textbf{Mixed non-spatial categorical attribute}:
    \begin{equation}
    c_{\text{mix}_i} = 
    \begin{cases} 
    c_i^s & \text{if } \mathcal{U}(0, 1) < \lambda, \\
    c_i^t & \text{otherwise.}
    \end{cases}
    \end{equation}
\end{enumerate}
,where the random variable $\lambda$ is drawn from a symmetric $\beta$ distribution with a range specified by the user-defined parameter $\alpha$. The non-spatial categorical attribute \(c_{\text{mix}_i}\) is determined based on a uniform distribution $\mathcal{U}$, such that \(c_{\text{mix}_i}\) is categorized as the source place-type if it is less than  $\lambda$; otherwise, it is categorized as the target place-type. To define our first self-supervised classification task, we consider instances from both the source and target place-types ($PT_S$ or $PT_T$) as positive samples, while instances derived from the mixed place-type $PT_{\text{mix}}$ are considered negative samples, as follows:
\begin{equation}
\begin{gathered}
\min_{\Phi_{\text{fea}}, \Phi_{cls_{mix}}} \mathcal{L}_{\text{cls}_{\text{mix}}} = -\frac{1}{n} \sum_{i=1}^{n} \left[ \mathit{I}_{\{x_i \in PT_S \cup PT_T\}} \log p_{i} \right. \\
\left. + \mathit{I}_{\{x_i \in PT_{\text{mix}}\}} \log (1 - p_{i}) \right],
\end{gathered}
\end{equation}
where $\mathit{I}_{\{x_i \in PT_S \cup PT_T\}}$ and $\mathit{I}_{\{x_i \in PT_{\text{mix}}\}}$ are indicator functions that equal 1 for samples belonging to \(PT_S\) or \(PT_T\) (positive samples) and \(PT_{\text{mix}}\) (negative samples), respectively, and \(p_i\) is the predicted probability of the \(i\)-th sample being positive.

\subsubsection{Spatial Masking}: A set of predefined geometries (e.g., square, rectangle) is used to mask the same sub-regions from pairs of source and target place-type instances. For each instance, the choice of geometry, the center location of each shape, and the proportion of masking in each batch are randomized to prevent models from overfitting to specific shapes and locations. Once the geometry is defined, all points within the mask from source $PT_S$ are replaced by those from target $PT_T$. The overall process of spatial masking and mixing produces a mask-mixed place type, denoted as $PT_{maskMix}$, along with the ratios of source and target instances in each instance of $PT_{maskMix}$. These ratios are detailed in a later part for a regression-
like self-supervised task.


The second self-supervised classification task involves discriminating among multiple classes. The first and second classes include instances from the source $0$ and target $1$ place-types, respectively. The third class $2$ includes instances that are mixed through local mask mixing perturbation. Given this set of local mask mix-up labels denoted by $L_{maskMix}$, and the total number of samples denoted by $N = n_S + n_T + n_{maskMix}$, the objective for our classifier is as follows:
\begin{equation}
\begin{gathered}
\min_{\Phi_{\text{fea}}, \Phi_{cls_{\text{maskMix}}}} 
\mathcal{L}_{cls_{\text{maskMix}}} = -\frac{1}
{N}\sum_{i=1}^{N}\sum_{l=1}^{L_{\text{maskMix}}} \\
\left[ \mathit{I}(l = PT_S) \log p_{i, l} \right] +  \left[ \mathit{I}(l = PT_T) \log p_{i, l} \right] + \\\left[ \mathit{I}(l = PT_{\text{maskMix}}) \log p_{i, l} \right].
\end{gathered}
\end{equation}
In the third self-supervised task, soft-labeling is used where the proportion of the preserved source instance in local mask mixing determines its soft-label, provided that the original label is known. The complementary label represents the proportion of the target instance within the revised mix-up sample. For instance, if the original source instance is labeled as $1$ and $80\%$ of it is preserved, the new soft-label for the source component becomes $0.8$ of the original label, and the remaining $0.2$ reflects the proportion of the target instance in $PT_{maskMisk}$. The objective function for this task is as follows:

\begin{equation}
\min_{\Phi_{\text{fea}}, \Phi_{cls_{\text{softMix}}}} \mathcal{L}_{cls_{\text{softMix}}} = -\frac{1}{N}\sum_{i=1}^{N} \left[\alpha_i \!\log p_{i,S} + \beta_i \!\log p_{i,T}\right],
\end{equation}

where $\alpha_i$ and $\beta_i$ are the proportions of the source and target instances in the $i$th mixed instance. For example, if 80\% of the mix comes from the source and 20\% from the target, then  $\alpha_i= 0.8$ and $\beta_i=0.2$. This approach introduces a more local spatial perturbation that encourages the model to adapt its understanding of source data by incorporating spatial arrangements from the target domain, thereby fostering the learning of domain-invariant features. More importantly, this method may also facilitate domain-specific insights, such as in oncology, by using spatial masking to simulate real-world spatial behaviors. For instance, spatial masking could mimic patterns of immune cell infiltration within tumor tissues, where specific regions are masked and replaced with cells from surrounding areas. This approach helps the model learn to recognize spatial variability between different types of tissue (e.g., healthy vs. tumor) while still leveraging the labels from the original source instances. 

\subsection{Spatial Contrastive Predictive Coding}: \label{SpatialCPC}Contrastive predictive coding \cite{oord2018representation} has shown great promise in unsupervised learning by predicting future states in latent space using powerful sequence models such as LSTM and auto-regressive models. However, unlike sequential models that inherently depend on time series, we employ an equivalent notion, that is the ``same place-type''. More precisely, our spatial contrastive predictive coding (SCPC) considers latent representations of instances from the same place-type as similar (i.e., positive pairs), and those from different place-types as dissimilar (i.e., negative pairs), as illustrated in the bottom right of Fig. \ref{Proposed}. 

Given \( PT_S = \{X_i^s\}_{i=1}^{n_s} \) and \( PT_T = \{X_i^t\}_{i=1}^{n_t} \) representing the input instances, and the corresponding latent encodings \((z_1^s, z_1^t, z_2^s, z_2^t, \dots, z_{n_s}^s, z_{n_t}^t)\) obtained from a DNN architecture (e.g., DGCNN \cite{wang2019dynamic}, SAMCNet\cite{farhadloo2022samcnet}), along with a context vector \( c_t \) from a recurrent model (e.g., GRU), then the objective is as follows:

\begin{equation}
\mathcal{L}_{\text{SCPC}} = -\frac{1}{N} \sum_{i=1}^{N} \log \frac{\exp(\text{sim}(\hat{\mathbf{z}}_j, \mathbf{z}_j) / \tau)}{\sum_{\mathbf{z}_k \in \{\mathbf{z}_j\} \cup \mathcal{N}(i)} \exp(\text{sim}(\hat{\mathbf{z}}_j, \mathbf{z}_k) / \tau)},    
\end{equation}
where \(\exp(\text{sim}(\hat{\mathbf{z}}_j, \mathbf{z}_j) / \tau)\) is the exponential of the similarity score for the positive pair (same place-type), scaled by the temperature parameter \(\tau\), \(\sum_{\mathbf{z}_k \in \{\mathbf{z}_j\} \cup \mathcal{N}(i)}\) is the summation over the positive pair \(\mathbf{z}_j\) and all negative pairs \(\mathbf{z}_k\) from the set \(\mathcal{N}(i)\), \(\text{sim}(\hat{\mathbf{z}}_j, \mathbf{z}_k)\) is the similarity score between the predicted latent vector \(\hat{\mathbf{z}}_j\) and a negative latent vector \(\mathbf{z}_k\), and \(-\frac{1}{N} \sum_{i=1}^{N}\) denotes averaging over all instances (i.e., \( N = \binom{n_s}{2} + \binom{n_t}{2} + n_s \times n_t \)).


\begin{table*}\scriptsize
\caption{Comparative analysis results on Classification Tasks via DGCNN \cite{wang2019dynamic} encoder}
\label{tab:my-table}
\centering
\renewcommand{\arraystretch}{0.9} 
\setlength{\tabcolsep}{2pt} 
\begin{adjustbox}{width=\linewidth} 
\begin{tabular}{|l|llll|llll|llll|llll|}
\hline
Task       & \multicolumn{4}{c|}{PT1ToPT2} & \multicolumn{4}{c|}{PT2ToPT1} & \multicolumn{4}{c|}{PT2ToPT3} & \multicolumn{4}{c|}{PT3ToPT2} \\ \hline
Method     & \multicolumn{1}{l|}{Accuray} & \multicolumn{1}{l|}{F1-score} & \multicolumn{1}{l|}{Precision} & Recall & \multicolumn{1}{l|}{Accuray} & \multicolumn{1}{l|}{F1-score} & \multicolumn{1}{l|}{Precision} & Recall & \multicolumn{1}{l|}{Accuray} & \multicolumn{1}{l|}{F1-score} & \multicolumn{1}{l|}{Precision} & Recall & \multicolumn{1}{l|}{Accuray} & \multicolumn{1}{l|}{F1-score} & \multicolumn{1}{l|}{Precision} & Recall \\ \hline
Supervised & 0.72 &0.72 &0.72 & 0.72 & 0.69 & 0.69 & 0.69 & 0.69 & 0.71 & 0.58 & 0.51 & 0.71 & 0.69 & 0.67 & 0.70 & 0.69 \\ \hline
{[}B1{] w/o adaptation}   & 0.46 & 0.46 & 0.46 & 0.46 & 0.57 & 0.41 & 0.32 & 0.57 &0.50 & 0.52 & 0.54 & 0.50 & 0.53 & 0.37 & 0.26 & 0.53 \\ \hline\hline
{[}B2{] DANN}   & 0.51 & 0.49 & 0.53  & 0.51 & 0.42  &  0.25 & 0.17 &  0.42 & 0.70 & 0.58   & 0.50 & 0.70  & 0.53  & 0.50 & 0.55  &  0.53  \\ \hline
{[}B3{] Rec + PCM}   & 0.48   & 0.31  & 0.23 & 0.48  & 0.60  & 0.45  & 0.36 & 0.60  &   0.70  & \textbf{0.64} & \textbf{0.66} & 0.70 & 0.53  & 0.36   & 0.28 & 0.53   \\ \hline
{[}B4{] GAST}   &  0.48   & 0.31  & 0.23 & 0.48 & 0.60  & 0.45  & 0.36 & 0.60  & 0.70 & 0.58  & 0.51 & 0.70 &0.53  & 0.38   & 0.51 & 0.53  \\ \hline
{[}P{] proposed} &\textbf{0.54} &\textbf{0.54}&\textbf{0.54}&\textbf{0.54}& \textbf{0.67} & \textbf{0.63} & \textbf{0.70} & \textbf{0.67} &  \textbf{0.71} & 0.58  & 0.51  & \textbf{0.71}  &  \textbf{0.61} & \textbf{0.58}  & \textbf{0.63} & \textbf{0.60}  \\ \hline
\end{tabular}
\end{adjustbox}
\label{cp_1}
\end{table*}

\begin{table*}\scriptsize
\caption{Comparative analysis results on Classification Tasks via SAMCNet \cite{farhadloo2022samcnet} encoder}
\label{tab:my-table}
\centering
\renewcommand{\arraystretch}{0.9} 
\setlength{\tabcolsep}{2pt} 
\begin{adjustbox}{width=\linewidth} 
\begin{tabular}{|l|llll|llll|llll|llll|}
\hline
Task       & \multicolumn{4}{c|}{PT1ToPT2} & \multicolumn{4}{c|}{PT2ToPT1} & \multicolumn{4}{c|}{PT2ToPT3} & \multicolumn{4}{c|}{PT3ToPT2} \\ \hline
Method     & \multicolumn{1}{l|}{Accuray} & \multicolumn{1}{l|}{F1-score} & \multicolumn{1}{l|}{Precision} & Recall & \multicolumn{1}{l|}{Accuray} & \multicolumn{1}{l|}{F1-score} & \multicolumn{1}{l|}{Precision} & Recall & \multicolumn{1}{l|}{Accuray} & \multicolumn{1}{l|}{F1-score} & \multicolumn{1}{l|}{Precision} & Recall & \multicolumn{1}{l|}{Accuray} & \multicolumn{1}{l|}{F1-score} & \multicolumn{1}{l|}{Precision} & Recall \\ \hline
Supervised & 0.93 & 0.94 & 0.94 & 0.94 & 0.85 & 0.85 & 0.85 & 0.85 & 0.93 & 0.93 & 0.94 & 0.94 & 0.92 & 0.91 & 0.91 & 0.91 \\ \hline
{[}B1{] w/o adaptation}   & 0.56 & 0.49 & 0.71 & 0.56 & 0.45 & 0.41 & 0.45 & 0.42 & 0.63 & 0.63 & 0.64 & 0.63 & 0.35 & 0.34 & 0.33 & 0.35 \\ \hline\hline
{[}B2{] DANN}   & 0.58 & 0.52 & 0.69 & 0.58  & 0.46 & 0.44 & 0.48 & 0.46  & 0.58      & 0.60   & 0.65  & 0.58  &  0.53  &  0.50   &   0.55    &  0.53  \\ \hline
{[}B3{] Rec + PCM}  & 0.48 & 0.45 & 0.48 & 0.48 & 0.58 & 0.43 & 0.34 & 0.58 &   0.63      & 0.63       & 0.67      & 0.63   &  0.54  &  0.51   &   0.53    &  0.54  \\ \hline
{[}B4{] GAST}   & 0.63 & \textbf{0.62} & \textbf{0.63}& 0.63 & 0.49 & 0.47 & 0.52 & 0.49 &  0.50      & 0.48        & 0.54      & 0.45 &  0.31  &  0.28   &   0.28    &  0.30  \\ \hline
{[}P{] proposed} &\textbf{0.67}&0.55 &0.55 &\textbf{0.67} & \textbf{0.82} & \textbf{0.84} & \textbf{0.86} & \textbf{0.84} &  \textbf{0.81} & \textbf{0.79}  & \textbf{0.83}  & \textbf{0.79}    & \textbf{0.68} & \textbf{0.67}  & \textbf{0.68} & \textbf{0.68} \\ \hline
\end{tabular}
\end{adjustbox}
\label{cp_2}
\end{table*}

\section{Validation} \label{sec:Experiment} 
\subsection{Experimental Settings: \label{setup}} Our experiments were designed to answer the following questions.
\begin{itemize}[noitemsep,topsep=0pt]
    \item [1.] How does proposed work compared to 
    competing DNN methods? 
    \item [2.]  How does the choice of deep learning architecture for learning spatial relationships affect classification performance?
    \item [3.]What impact do spatial self-supervised tasks have on solution quality?
\end{itemize}

\textbf{Datasets:} The experiments were conducted using a real-world cancer dataset derived from MxIF images \footnote{Due to patient privacy, the dataset is not shared, but the experiment code is available \href{https://drive.google.com/drive/folders/1m4QH5hKhXqhQ11Hg1qtfbF98UZV54HnU}{here}.}. This dataset consists of three distinct place-types: (1) \textbf{normal} denoted as \textbf{$PT_1$}, (2) \textbf{interface}, denoted as \textbf{$PT_2$}, and (3) \textbf{tumor}, denoted as \textbf{$PT_3$}. For place-type \textbf{$PT_1$}, the dataset contained 81 multi-type point maps representing two different clinical outcomes of immune therapy. Of these, 38 sets were labeled as responders, while 43 were classified as non-responders, signifying individuals who progressed and experienced tumor recurrence within a year. For place-type \textbf{$PT_2$}, the dataset contained 145 multi-type point maps. Out of these, 68 were identified as responders, with the remaining 77 labeled as non-responders. Lastly, in place-type \textbf{$PT_3$}, out of the 103 point sets provided, 30 were labeled as responders, and 73 as non-responders. The first classification task ($\textbf{PT1ToPT2}$) involved training on source place-type ($PT_1$) and testing on target place-type ($PT_2$), while the second task ($\textbf{PT2ToPT1}$) operated in the opposite direction. The third classification task ($\textbf{PT2ToPT3}$) entailed training on a source place-type ($PT_2$) and testing on target place-type ($PT_3$), with the fourth task ($\textbf{PT3ToPT2}$) reversing this direction.

\textbf{Dataset Preparation:} In each classification task, we divided the data into 80\% training and 20\% testing. Twenty five percent of the training set was selected to be the validation set. Due to the limited number of learning samples, we used the farthest point sampling strategy to break each instance into uniform subsets of 1,024 points for each classification task.

\textbf{Deep Learning  Architectures:} We compared our proposed framework on selected classification metrics with the following state-of-the-art DNN architectures as the backbone for the encoder $\Phi$: \textbf{(1) DGCNN} \cite{wang2019dynamic}, a dynamic graph convolutional neural network architecture for CNN-based high-level point cloud tasks such as classification and segmentation; and \textbf{(2) SAMCNet}, a spatial-interaction aware multi-category deep neural network \cite{farhadloo2022samcnet}, for learning N-way spatial relationships in multi-type point maps. All hyper-parameters were tuned through tuning on the validation set. 

\textbf{Evaluation Metric \& Platform:} Model performance was assessed using the weighted average of accuracy, precision, recall, and F1-score. This ``weighted'' method accounts for class imbalance by calculating the average of these binary metrics, where the contribution of each class’s score is weighted according to its prevalence in the true data sample \cite{scikit-learn}. We used a K40 GPU with 40 Haswell Xeon E5-2680 v3 nodes. Each node had 128 GB of RAM and NVidia Tesla K40m GPUs, each with 11 GB of RAM and 2880 CUDA cores.

\textbf{Candidate Methods:} We evaluated our approach by testing its performance in two scenarios: one where the model was trained using supervised classification on the target place-type, and another where it was trained on the source without domain adaptation. We compared our method to state-of-the-art unsupervised point-based domain adaptation techniques, including DANN \cite{ganin2016domain}, DefRec+PCM \cite{achituve2021self}, and GAST \cite{zou2021geometry}. These well-established baselines are widely used in transfer learning for domain adaptation and have been applied in similar studies \cite{achituve2021self, shen2022domain, zou2021geometry, Liu2023} addressing cross-domain classification challenges. This ensures that our findings are framed within the broader research landscape.

\subsection{Experiment Results}: Results of our spatially-delineated domain-adapted AI classification assessment were as follows.

\emph{\textbf{Comparative Analysis:}}
We conducted experiments to evaluate various DNN architectures for the classification tasks described in Section \ref{setup}. The results across all tasks are summarized in Tables \ref{cp_1} and \ref{cp_2}. These results demonstrated the superiority of our proposed method over the existing DNN competition, including DANN, DefRec+PCM, and GAST, across most of the classification metrics. Most notably, as shown in Table \ref{cp_1}, our method using DGCNN as the encoder improved classification accuracy over the best competitor by margins of 6.0\%, 7.0\%, and 8.0\% in the $PT1ToPT2$, $PT2ToPT1$, and $PT3ToPT2$ classifications, respectively. Similarly, our method with SAMCNet significantly enhanced classification accuracy over the best competitor by margins of 4\%, 24\%, 18.0\%, and 14.0\% in the $PT1ToPT2$, $PT2ToPT1$, $PT2ToPT3$, and $PT3ToPT2$ classifications, respectively, as illustrated in Table \ref{cp_2}. More importantly, our method consistently yielded positive adaptation gains, whereas the existing competition often resulted in negative or no adaptation gains compared to the lower bound. 
\begin{table*}[h]\scriptsize
\caption{Sensitivity analysis results on Classification Tasks}
\label{tab:my-table}
\centering
\renewcommand{\arraystretch}{0.9} 
\setlength{\tabcolsep}{2pt} 
\begin{adjustbox}{width=\linewidth} 
\begin{tabular}{|l|llll|llll|llll|llll|}
\hline
Task       & \multicolumn{4}{c|}{PT1ToPT2} & \multicolumn{4}{c|}{PT2ToPT1} & \multicolumn{4}{c|}{PT2ToPT3} & \multicolumn{4}{c|}{PT3ToPT2} \\ \hline
Method     & \multicolumn{1}{l|}{Accuray} & \multicolumn{1}{l|}{F1-score} & \multicolumn{1}{l|}{Precision} & Recall & \multicolumn{1}{l|}{Accuray} & \multicolumn{1}{l|}{F1-score} & \multicolumn{1}{l|}{Precision} & Recall & \multicolumn{1}{l|}{Accuray} & \multicolumn{1}{l|}{F1-score} & \multicolumn{1}{l|}{Precision} & Recall & \multicolumn{1}{l|}{Accuray} & \multicolumn{1}{l|}{F1-score} & \multicolumn{1}{l|}{Precision} & Recall \\ \hline
{SMUM}& 0.63 & 0.58 & 0.71 & 0.63&0.63 & 0.58 & 0.71 & 0.63 & 0.67 & 0.67 & .67 & 0.67 & 0.64 & 0.60 & 0.68 & 0.64   \\ \hline 
{SCPC} & 0.58 & 0.58 & 0.67 & 0.59 & 0.63 & 0.63 & 0.64 & 0.63 & 0.79 & 0.77 & 0.79 & 0.79 & 0.66 & 0.67 & 0.67 & 0.66 \\ \hline
\end{tabular}
\end{adjustbox}
\label{sens1}
\end{table*}

From a clinical perspective, $PT2$ serves as an intermediate representation between the periphery ($PT1$) and the core ($PT3$), reflecting spatial transitions between these regions. This positioning likely facilitates closer alignment with supervised settings when transitioning to $PT1$ and $PT3$. Conversely, the greater heterogeneity and class imbalances in $PT1$ and $PT3$—compared to the more intermediate $PT2$—may hinder the model's ability to effectively learn and transfer knowledge to $PT2$.

Moreover, it can be observed that the choice of DNN architecture may play a significant role due to the importance of learning spatial relationships in multi-type point maps. Our proposed
with SAMCNet as encoder was able to improve accuracy over DGCNN \cite{wang2019dynamic} by a margin of 13\%, 15\%, 6\%, and 7\% in the $PT1ToPT2$, $PT2ToPT1$, $PT2ToPT3$, and $PT3ToPT2$ classifications, respectively.

\emph{\textbf{Sensitivity Analysis}}: To evaluate the primary self-supervised learning tasks in our proposed framework, we explored how the model performs with and without these tasks. We focused on key elements like spatial mix-up masking (SMUM) and spatial contrastive predictive coding (SCPC), and we used a variety of classification measures to assess their impact. The findings are presented in Table \ref{sens1}. SAMCNet \cite{farhadloo2022samcnet} served as the backbone encoder $\Phi$ for this experiment.

The results show that using SMUM (Spatial Mix-Up Masking) is beneficial in exposing the model to a diverse set of spatial arrangements and facilitating domain-specific insights, where spatial masking can potentially mimic patterns such as immune cell infiltration within tumor tissues. Similarly, SCPC (Spatial Context Prediction Consistency) is effective in learning spatial arrangements that are consistent within each place-type while distinguishing between different place-types. By applying an autoregressive model that predicts the latent embedding of another instance from the same place-type, SCPC encourages the model to capture meaningful spatial dependencies within each place-type. Yet, the proposed model performs best when both submodules—SMUM and SCPC—are integrated and work together as a cohesive unit.

\section{Case Study}\label{case-study}
We conducted a case study aimed at comparing intracellular interactions using our proposed framework versus a supervised learning method on the target place-type, to classify input samples as either responder or non-responder. For this, we utilized a trained SAMCNet \cite{farhadloo2022samcnet} DNN to extract features after the point pair prioritization network at layer-4, where the model has learned spatial and prioritization associations.
We evaluated the significance of the identified spatial relationships using permutation feature importance. This metric measures the importance of a feature by assessing the decrease in model performance when the feature space is randomly shuffled. Previous studies show the effectiveness of interpretable models that employ hand-constructed spatial quantification methods (e.g., participation ratio \cite{shekhar2001discovering}), combined with decision tree algorithms \cite{li2021srnet}. Tables \ref{core1} and \ref{core2} list the top five most relevant spatial associations found within tumor-core $PT_3$, and a supervised method trained solely on the target place-type samples. The following provides a brief interpretation of these results from a clinical perspective.

\textbf{Clinical Implications:} 
What is known to correlate with clinical outcomes is the penetration of cytotoxic T lymphocyte cells (CD8) into tumor-core. Thus, the presence of more tumor-infiltrating CD8 cells typically indicates better outcomes. This is because CD8 immune cells are the only ones capable of destroying cancer cells, and they must physically contact these cells to do so. However, not all CD8 cells can kill tumor cells; for example, CD8 immune cells with PD1 receptors can be inhibited by PDL1 ligands (found on tumors or cells surrounding tumors like tumor-infiltrating macrophages), which disable the CD8 cells' ability to kill. Therefore, the effectiveness of an infiltrating CD8 cell is heavily influenced by its immediate environment and the cells surrounding the cancer target cell. This suggests a dynamic relationship involving a target cell (tumor cell), an effector cell (CD8 T cell), and a modulator cell (nearby cell regulating interactions between the target and effector cells).


\begin{table*}
\centering
\caption{The most relevant spatial relationships in the tumor core place using the proposed approach.}
\begin{tabular}{|c|c|}
\hline
\textbf{Rank} & \textbf{Cell arrangements} \\ \hline
1 & \textless Tumor cell\textgreater \textless Macrophage, Tumor cell, Vasculature\textgreater \\ \hline
2 & \textless Tumor cell\textgreater \textless Macrophage, Tumor cell\textgreater \\ \hline
3 & \textless Tumor cell\textgreater \textless Cytotoxic T cell, Macrophage, Regulatory T cell, Tumor cell\textgreater \\ \hline
4 & \textless Regulatory T cell\textgreater \textless Macrophage, Regulatory T cell, Tumor cell, Vasculature\textgreater \\ \hline
5 & \textless Neutrophil\textgreater \textless Neutrophil, Tumor cell\textgreater \\ \hline
\end{tabular}
\label{core1}
\end{table*}

Our proposed method demonstrates that factors beyond the ``tumor-cell/CD8 cell'' interaction are relevant, and it suggests that ``macrophages'' may serve as the ``modulator'' cell. Therefore, it is crucial to explore how macrophages can interfere with the CD8/tumor cell interaction and influence the survival of the tumor cell. Macrophages may affect this interaction either through secreted factors that influence both cells as they interact or through molecules expressed on their surface that modulate CD8 cell activity towards the tumor cell. One such model involves the expression of PDL1 on macrophages; PDL1 on the macrophage directly engages the CD8 PD1 receptor to deactivate the CD8 cell, rendering it incapable of killing the tumor cell. However, this is just one of many mechanisms, all of which require close contact between the 'modulator' cell and the CD8-tumor cell complex. Additionally, this cellular arrangement can also be observed through a supervised method, which confirms that proposed method learns and preserves critical spatial cell interactions.

\begin{table*}
\centering
\caption{The most relevant spatial relationships in the tumor core place using the supervised approach.}
\begin{tabular}{|c|c|}
\hline
\textbf{Rank} & \textbf{Cell arrangements} \\ \hline
1 & \textless Vasculature\textgreater \textless Macrophage, Tumor cell\textgreater \\ \hline
2 & \textless Tumor cell\textgreater \textless Macrophage, Tumor cell\textgreater \\ \hline
3 & \textless Tumor cell\textgreater \textless Cytotoxic T cell, Macrophage, Regulatory T cell, Tumor cell\textgreater \\ \hline
4 & \textless Regulatory T cell\textgreater \textless Macrophage, Regulatory T cell, Tumor cell\textgreater \\ \hline
5 & \textless Tumor cell\textgreater \textless Cytotoxic T cell, Tumor cell, Vasculature\textgreater \\ \hline
\end{tabular}
\label{core2}
\end{table*}

\section{Related Work}\label{related_work}
Domain adaptation (DA) addresses the challenge of utilizing a well-labeled source domain dataset to enhance performance on a completely unlabeled target domain dataset. Most prior research in unsupervised domain adaptation for spatially-delineated classification of multi-type point maps is classified into adversarial \cite{ganin2016domain, NIPS2019_8940, debortoli2021adversarial, saito2018maximum} or self-supervised learning approaches \cite{achituve2021self, shen2022domain, zou2021geometry, Liu2023}.

\textbf{Unsupervised Adversarial Adaptation}: Adversarial techniques aim to align source and target domains using a min-max optimization strategy between discriminators and generators \cite{goodfellow2020generative}. The idea is that the feature representation should be good for the main learning task while being similar enough between the two domains \cite{ganin2016domain, saito2018maximum}. Early work includes the domain adversarial neural network \cite{ganin2016domain}, which uses a gradient reversal layer to mitigate distribution shifts and ensure similar feature distributions across the two domains. Adversarial discriminative domain adaptation generalizes domain adaptation by independently mapping source and target domains with untied weights, initializing the target model from a pre-trained source model, and fine-tuning it with adversarial training \cite{long2018conditional}.

These frameworks have been adapted for point set domain adaptation. PointDAN\cite{NIPS2019_8940}, the first work to address point set classification via domain adaptation, used adversarial training with maximum classifier discrepancy along with GRL to jointly learn and align both local and global features across source and target domains \cite{NIPS2019_8940, saito2018maximum}. PointDGAN extends PointDAN by integrating generative adversarial networks (GANs) to generate synthetic point clouds that bridge the gap between source and target domains \cite{huang2022generation}. Despite fundamental works and significant advancements, adversarial training can still lead to degenerated local minima, potentially resulting in negative adaptation gains.

\textbf{Self-supervised Learning  Adaptation}: Self-supervised learning techniques focus on designing tasks that do not require labeled data but instead leverage the inherent structure of the data to learn domain-invariant features and mitigate domain differences. Achituve et al. \cite{achituve2021self} enhanced model training by introducing volume-based and sample-based deformation reconstruction for shape prediction, and point cloud mix-up to estimate proportions of mixed inputs. GAST \cite{zou2021geometry} has introduced a geometry-aware, self-supervised training method that encodes domain-invariant geometric features into semantic representations to mitigate domain discrepancies in point-based representations. Shen et al. \cite{shen2022domain} leverage implicit representations to capture and preserve the underlying geometry of point sets. The core idea is to align the implicit representations across different domains, moving beyond mere point-wise feature alignment to mitigate domain shifts through geometry-aware implicit representations. 

While state-of-the-art methods, all these works overlook the underlying spatial arrangements among data points, leading to substantial discrepancies among different place types. Moreover, similar works \cite{he2022masked, zhou2021domain, zhang2017mixup, yu2022point} have explored feature alignments between source and target instances in a shared latent space across various data modalities (e.g., images, feature vectors) to mitigate distribution shifts between domains. Our work leverages ideas from these studies to explicitly target both global and local underlying spatial arrangements that respect the inherent structure of the data for domain adaptation across place-types while preserving our understanding of crucial spatial patterns.

\section{Conclusion \& Future Work} \label{sec:Conclusion}
We investigated a spatially-delineated domain adapted classification DNN for multi-type point maps. Our approach introduces a multi-task learning framework where surrogate classification tasks through self-supervised tasks are defined to learn representations that capture the underlying spatial arrangements across different place-types. Experiments show that the proposed model outperforms existing DNN techniques.

For future research, we aim to integrate domain-specific knowledge into self-supervised learning tasks to better capture spatial relationships across place-types. We also aim to examine the heterogeneity and transitional patterns between source and target place types, focusing on bi-directional domain adaptation strategies to optimize mutual information \cite{french2018selfensembling, li2019bidirectional}. Additionally, while numerous models have been proposed, limited attention has been given to the data itself. Thus, we plan to investigate spatially-characterized synthetic data generation methods that preserve the structural characteristics of the original data while ensuring privacy. This approach will enable more controlled experimentation and enhance our understanding of spatial variability.

\section*{Acknowledgments}{This material is based on work supported by the NSF under Grants Nos. 1901099, and 1916518; and the USDA under Grant Nos. 2023-67021-39829 and 2021-51181-35861. We also thank Kim Koffolt and Spatial Computing Research Group for their valuable comments and refinements.}

\bibliographystyle{unsrt}
\bibliography{citation}

\end{document}